\documentclass{llncs}  
\usepackage{include/llncsdoc}
\usepackage{multicol}
\usepackage[bookmarks=true]{hyperref}
\usepackage{amsmath}
\usepackage{textcomp}
\usepackage{graphicx}
\usepackage[font=scriptsize]{caption}
\usepackage{hyperref}
\usepackage{amssymb}
\usepackage{booktabs}
\usepackage[normalem]{ulem}
\usepackage{verbatim}
\usepackage[export]{adjustbox}
\usepackage{url}
\usepackage{siunitx}
\usepackage[utf8]{inputenc}
\usepackage[TS1,T1]{fontenc}
\usepackage{array, booktabs}
\usepackage{float}
\usepackage{bm}
\usepackage{multirow}
\usepackage{url}
\usepackage{amssymb,amsfonts,amsfonts}
\usepackage{algorithm}
\usepackage[noend]{algorithmic}
\usepackage{xspace}
\usepackage{wrapfig}
\usepackage{color}
\usepackage{hyperref}
\let\labelindent\relax
\usepackage{enumitem}
\usepackage{times}
\usepackage{helvet}
\usepackage{courier}
\usepackage[title]{appendix}
\definecolor{Gray}{gray}{0.5}
\usepackage[rightcaption]{sidecap}

\newcommand{\eref}[1]{(\ref{#1})} 
\newcommand{\figref}[1]{Fig.~\ref{#1}} 

\usepackage{blindtext}
\usepackage{ifthen}
\usepackage[usenames,dvipsnames]{xcolor}

\usepackage{subfigure}

\usepackage{array}
\usepackage{tabularx}
\usepackage{multirow}
\usepackage{multicol}
\usepackage{booktabs}
\usepackage{epstopdf}
\usepackage{rotating}
\usepackage{csquotes}
\usepackage[bookmarks=true]{hyperref}

\usepackage[export]{adjustbox}
\usepackage{amssymb}
\usepackage{amsmath}
\usepackage[all]{xy}    
\usepackage{ifthen}
\usepackage{colortbl}
\usepackage{enumerate}
\usepackage{bm}
\usepackage{hhline}

\usepackage{algorithm}
\usepackage{algorithmic}
  \usepackage{setspace}
\let\Algorithm\algorithm
\renewcommand\algorithm[1][]{\Algorithm[#1]\setstretch{1.2}}

\newcommand{\secref}[1]{Section~\ref{#1}}
\newcommand{\tabref}[1]{Table~\ref{#1}}
\newcommand{\algref}[1]{Algorithm~\ref{#1}}

\newcommand{\myparagraph}[1]{\vspace{0.1in}\noindent\textbf{#1}}

\begin{document}
\mainmatter              
\title{GP-SUM. Gaussian Process Filtering of non-Gaussian Beliefs}
\titlerunning{GP-SUM}  
%
\author{Maria Bauza and Alberto Rodriguez}
\authorrunning{Bauza and Rodriguez} 
%
\tocauthor{Maria Bauza and Alberto Rodriguez}
\institute{Mechanical Engineering Department --- Massachusetts Institute of Technology\\
\email{<bauza,albertor>@mit.edu} }

\maketitle              

\begin{abstract}
This work studies the problem of stochastic dynamic filtering and state propagation with complex beliefs.
The main contribution is GP-SUM, a filtering algorithm tailored to dynamic systems and observation models expressed as Gaussian Processes (GP), and to states represented as a weighted \emph{Sum} of Gaussians.
The key attribute of GP-SUM is that it does not rely on linearizations of the dynamic or observation models, or on unimodal Gaussian approximations of the belief, hence enables tracking complex state distributions.

The algorithm can be seen as a combination of a sampling-based filter with a probabilistic Bayes filter. On the one hand, GP-SUM operates by sampling the state distribution and propagating each sample through the dynamic system and observation models.
On the other hand, it achieves effective sampling and accurate probabilistic propagation by relying on the GP form of the system, and the sum-of-Gaussian form of the belief.

We show that GP-SUM outperforms several GP-Bayes and Particle Filters on a standard benchmark. We also demonstrate its use in a pushing task, predicting with experimental accuracy the naturally occurring non-Gaussian distributions.

  %
%
\end{abstract}

\section{Introduction}
\label{sec:introduction}
	
Robotics and uncertainty come hand in hand. 
One of the defining challenges of robotics research is to design uncertainty-resilient behavior to overcome noise in sensing, actuation and/or dynamics.
This paper studies the problems of simulation and filtering in systems with stochastic dynamics and noisy observations, with a particular interest in cases where the state belief cannot be realistically approximated by a single Gaussian distribution. 

\begin{figure}[t]
\centering
\subfigure{\includegraphics[width=2.2in]{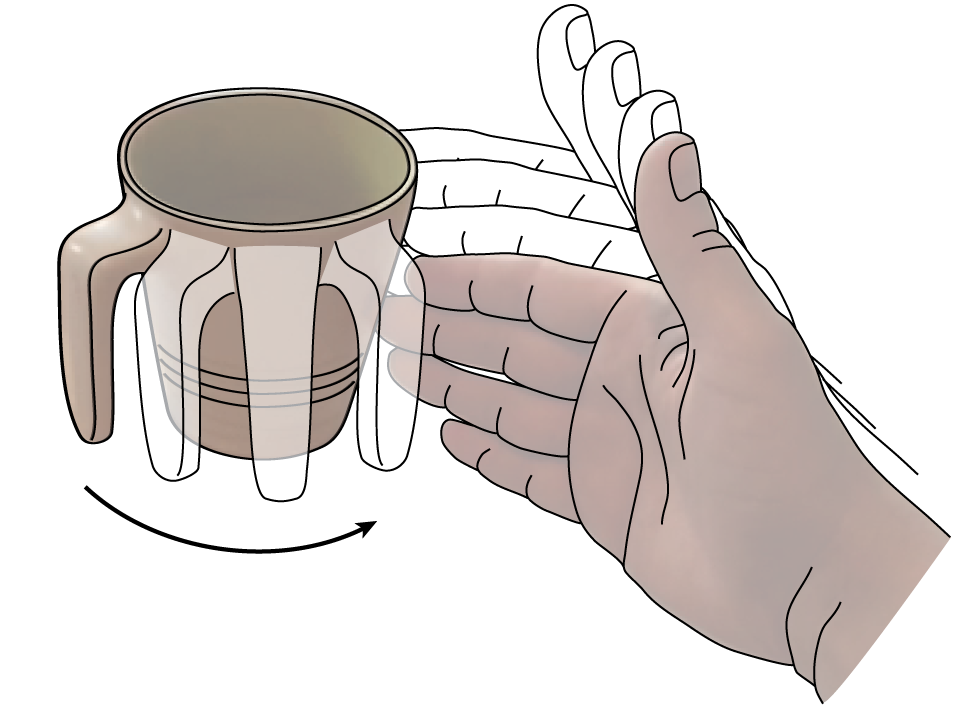}} 
\subfigure{\includegraphics[width=2.2in]{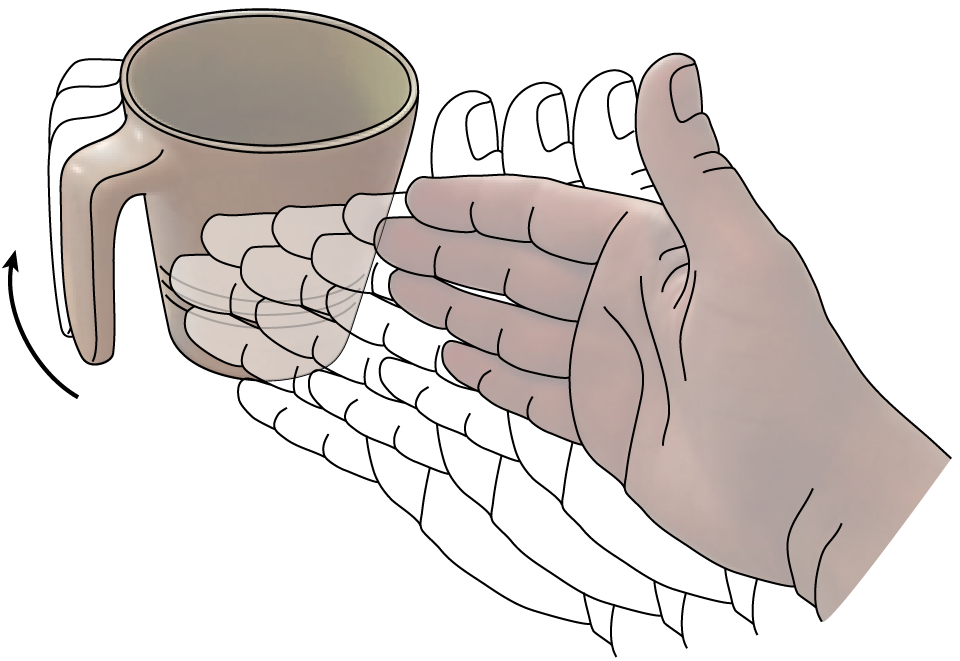}}
\caption{\textbf{Multimodality during the push-grasp of a coffee cup.} A small change in the initial contact between the hand and the cup, or a small change in the hand's motion can produce a very distinct---multimodal---outcome either exposing (left) or hiding (right) the cup's handle from the hand's palm.}
\label{fig:grasp_mug}
\end{figure}

Complex multimodal beliefs naturally arise in manipulation tasks where state or action noise can make the difference between contact/separation or between sticking/sliding.
For example, the ordinary task of push-grasping a cup of coffee into your hand in \figref{fig:grasp_mug} illustrates the naturally occurring multimodality, where the cup will slide to either side of the hand and the handle will rotate accordingly clockwise or anti-clockwise.
Dogar and Srinivasa \cite{Dogar2010} used the observation that a push-grasped object tends to cluster into two clearly distinct outcomes---inside and outside the hand---to plan robust grasp strategies. 
Similarly, as illustrated in \figref{fig:push_dist}, a push on an object might make it rotate to either side of the pusher, generating complex---clustered and ring-shaped---state distributions.
Multimodality and complex belief distributions have been experimentally observed in a variety of manipulation actions such as planar pushing~\cite{Yu2016,Bauza2017}, ground impacts~\cite{Fazeli2017b}, and bin-picking~\cite{Paolini2014}.

%

The main contribution of this paper is a new algorithm GP-SUM to track complex state beliefs. GP-SUM is specifically tailored to:
\begin{itemize}
    \item Dynamic systems expressed as a \emph{GP} (Gaussian Process).
    \item States represented as a weighted \emph{Sum} of Gaussians.
\end{itemize}
We will show that the algorithm is capable of performing Bayes updates without the need to either linearize the dynamic or observation models, or relying on unimodal Gaussian approximations of the belief. This will be key to enable efficient tracking of complex state beliefs.

In \secref{sec:mixture-filter} we describe how GP-SUM operates by sampling the state distribution, given by a sum of Guassians, so it can be viewed as a sampling-based filter. GP-SUM also maintains the basic structure of a Bayes filter by exploiting the GP form of the dynamic and observation models, which allows a probabilistic sound interpretation of each sample, so it can also be viewed as a GP-Bayes filter. 


In \secref{sec:comparison}, we compare GP-SUM's performance to other existing GP-filtering algorithms such as GP-UKF, GP-ADF and GP-PF in a standard synthetic benchmark~\cite{Deisenroth2009,Pan2017}. 
GP-SUM yields better filtering results both after a single and multiple filtering steps with a variety of metrics, and requires significantly less samples than standard particle filtering techniques.

Finally, we also demonstrate that GP-SUM can  predict the expected distribution of outcomes when pushing an object. Prior experimental work~\cite{Bauza2017} shows that planar pushing produces heteroscedastic and non-Gaussian distribution after pushes of a few centimeters, i.e., some actions are more deterministic than others and state distributions can break down into components or become ring-shaped. 
GP-SUM successfully recovers both when applied to a GP learned model of planar pushing. We compare the results to the distributions from real trajectories estimated by performing the same pushes 100 times.

Both actions and sensed information determine the shape of a belief distribution and non-linearities in either when integrated over a finite time can easily lead to non-Gaussian beliefs. This paper provides an efficient algorithm for tracking complex distributions tailored to the case where the observation and transition models are expressed as GPs, and the state is represented as a weighted sum of Gaussians. 

%



\section{Related Work}
\label{sec:relatedWork}

Gaussian Processes (GPs) are a powerful tool to model the dynamics of complex systems \cite{Paolini2014,Kocijan2005,Nguyen2009}, and have been applied to different aspects of robotics including planning and control~\cite{Murray2002,Deisenroth2012,Mukadam2016}, system identification~\cite{Kocijan2005,Gregorvcivc2008,Avzman2011}, or filtering~\cite{Deisenroth2009,Pan2017,Ko2009}. 
In this work we study the problem of accurate propagation and filtering of the state of a stochastic dynamic system. In particular, we address filtering in systems whose dynamics and measurement models are learned through GP regression, which we commonly refer to as GP-Bayes filters. Among these, the most frequently considered are GP-EKF \cite{Ko2009}, GP-UKF \cite{Ko2009} and GP-ADF \cite{Deisenroth2009}.

Most GP-filters rely on the assumption that at all instants, the state distribution is well captured by a single Gaussian and exploit a variety of approximations to maintain that Gaussianity. 
For example, GP-EKF is based on the extended Kalman filter (EKF) and linearizes the GP models to guarantee that the final distributions are indeed Gaussian. GP-UKF is based on the unscented Kalman filter (UKF) and predicts a Gaussian distribution for the state using an appropriate set of samples that captures the moments of the state. Finally, GP-ADF computes the first two moments of the state distribution by exploiting the structure of GPs and thus returns a Gaussian distribution for the state.

GP-SUM instead is based on sampling from the state distributions and using Gaussian mixtures to represent these probabilities. This links our algorithm to the classical problem of particle filtering where each element of the mixture can be seen as a sample with an associated weight and a Gaussian. As a result, GP-SUM can be understood as a type of sampling algorithm that is tailored to exploit the synergies between a GP-based dynamic model and a Gaussian mixture state to enable efficient and probabilistically-sound Bayes updates. 
Ko and Fox~\cite{Ko2009} provide another GP-based sampling filter, GP-PF, based on the classical particle filter. However, when compared to GP-UKF or GP-EKF, GP-PF is less reliable and more prone to inconsistent results~\cite{Ko2009}. 

In the broader context of Bayes filtering with non-linear algebraic dynamic and observation models, multiple algorithms have been proposed to recover non-Gaussian state distributions. For instance, there is some resemblance between GP-SUM and the algorithms Gaussian Mixture Filter (GMF)~\cite{Stordal2011}, Gaussian Sum Filter (GSF)~\cite{alspach1972}, and Gaussian Sum Particle Filtering (GSPM)~\cite{Kotecha2003}; all using different techniques to propagate the state distributions as a sum of Gaussians.
GMF considers a Gaussian mixture model to represent the state distribution, but the covariance of all Gaussians are equal and come from sampling the previous state distribution and computing the covariance of the resulting samples; GP-SUM instead computes directly the covariance of the mixture from the GP system dynamics.
GSF is as a set of weighted EKFs running in parallel. As a consequence it requires linearization of the system dynamics and observation models while GP-SUM does not. 
Finally GSPM, which has proved to outperform GSF, is based on the sequential importance sampling filter (SIS) ~\cite{Kotecha2003}. GSPM takes samples from the importance function which is defined as the likelihood of a state $x$  given an observation z, $p(x|z)$. GP-SUM instead does not need to learn this extra mapping, $p(x|z)$, to effectively propagate the state distributions.

Other more task-specific algorithms also relevant to GP-SUM are the multi-hypothesis tracking filter (MHT)~\cite{Blackman2004} and the manifold particle filter (MPF)~\cite{Koval2017}. 
MHT is designed to solve a data association problem for multiple target tracking by representing the joint distribution of the targets as a Gaussian mixture. 
MPF is a sample-based algorithm taylored to dynamic systems involving unilateral contact constraints which induce a decomposition of the state space into subsets of different dimension, e.g., free space versus contact space. MPF exploits an explicit model of the contact manifolds of the system to project the distributions defined by the samples into that manifold. 

An advantage of GP-SUM is that it can be viewed as both a sampling technique and a parametric filter. Therefore most of the techniques employed for particle filtering are applicable. Similarly, GP-SUM can also be adapted to special types of GPs such as heteroscedastic or sparse GPs. For instance, GP-SUM can be easily combined with sparse spectrum Gaussian processes (SSGPs) in Pan et al.~\cite{Pan2017}. Consequently, the learning, propagation and filtering of the dynamics can be made significantly faster.


\section{Background on Gaussian process filtering}
\label{sec:problem}

This work focuses on the classical problem of Bayes filtering where the dynamics and observation models  are learned through Gaussian process regression. In this section, we introduce the reader to the concepts of Bayes filtering and Gaussian processes. 

\subsection{Bayes filters}

The goal of a Bayes filter is to track the state of a system, $x_t$, in a probabilistic setting.  At time $t$, we consider that an action $u_{t-1}$ is applied to the system making its state evolve from $x_{t-1}$ to $x_t$. This is followed by an observation of the state,  $z_{t}$. As a result, a Bayes filter computes the state distribution, $p(x_t)$, conditioned on the history of previous actions and observations: $p(x_t | u_{1:t-1}, z_{1:t})$. This probability is often referred as the belief of the state at time $t$.

In general, a Bayes filter is composed of two steps: the prediction update and the measurement or filter update following the terminology from \cite{Thrun2005}.

\myparagraph{Prediction update.} 
Given a model of the system dynamics, $ p(x_t | x_{t-1}, u_{t-1})$, the prediction update computes the \textit{prediction belief}, $p(x_t| u_{1:t-1}, z_{1:t-1})$, as:
\begin{equation}
\begin{gathered} 
  p(x_t | u_{1:t-1}, z_{1:t-1}) = \int  p(x_t | x_{t-1}, u_{t-1}) p(x_{t-1} | u_{1:t-2}, z_{1:t-1})d x_{t-1}  
\label{eq:prediction_believe} 
\end{gathered}
\end{equation}
where $p(x_{t-1} | u_{1:t-2}, z_{1:t-1})$ is the belief of the system before action $u_{t-1}$. Thus the prediction belief can be understood as the \textit{pre-observation} distribution of the state, while the belief is the \textit{post-observation} distribution. In general, the integral \eref{eq:prediction_believe} cannot be solved analytically and different approximations are used to simplify its computation. The most common simplifications are to linearize the dynamics of the system, as classically done in the Extended Kalman Filter, or to directly assume that the prediction belief, i.e., the result of the integral in \eref{eq:prediction_believe}, is Gaussian distributed \cite{Thrun2005}.

\myparagraph{Measurement update.} 
Given a new measurement of the state, $z_t$, the belief at time $t$ comes from filtering the prediction belief. The belief is recovered by using Bayes' rule and the observation model of the system $p(z_t| x_t)$:
\begin{equation}\label{eq:belief}
p(x_t | u_{1:t-1}, z_{1:t})  = \frac{p(z_t| x_t)p(x_t | u_{1:t-1}, z_{1:t-1})}{p(z_t| u_{1:t-1}, z_{1:t-1})}
\end{equation}
Again, this expression cannot usually be computed in a closed-form and we rely on approximations to estimate the new belief. Linearizing the observation model or assuming Gaussianity of the belief are again common approaches~\cite{Thrun2005}.

Combining equations \eref{eq:prediction_believe} and \eref{eq:belief}, we can express the belief in a recursive manner as a function of the previous belief, the dynamic model, and the observation model:
\begin{equation}
\begin{gathered}
\label{eq:belief_v2} 
p(x_t | u_{1:t-1}, z_{1:t}) \propto p(z_t| x_t) \int  p(x_t | x_{t-1}, u_{t-1}) p(x_{t-1}| u_{1:t-2}, z_{1:t-1}) d x_{t-1}
\end{gathered}
\end{equation}
We will show in \secref{sec:mixture-filter} an equivalent recursion equation for the prediction belief, which is key to GP-SUM.

For known systems, we might have algebraic expressions for their dynamic and observation models. In real systems, however, these models are often unknown or inaccurate, and Gaussian Processes are a powerful framework to learn them. The following subsection provides a basic introduction.

\subsection{Gaussian Processes}
Gaussian Processes (GPs) are a flexible non-parametric framework for function approximation~\cite{Rasmussen2006}. In this paper we use GPs to model the dynamics and observation models of a stochastic system. 
There are several advantages from using GPs over traditional parametric models. First, GPs can learn high fidelity models from noisy data while estimate the intrinsic noise of the system. Second, GPs estimate the uncertainty of their predictions given the available data, hence measuring the quality of the regression. GPs provide the value of the expected output together with its variance. 

In classical GPs~\cite{Rasmussen2006}, the noise in the output is assumed to be Gaussian and constant over the input:  $y(x) = f(x) + \varepsilon $,  where $f(x)$ is the latent or unobserved function to regress, $y(x)$ is a noisy observation of this function,  and $\varepsilon \sim N(0, \sigma^2)$ represents zero-mean Gaussian noise with constant variance $\sigma^2$.

The assumption of constant Gaussian noise together with a GP prior on the latent function $f(x)$ makes analytical inference possible for GPs. In practice, to learn a GP model over $f(x)$ you only need a set of training points, $D = {\{(x_i, y_i)\}_{i=1}^n}$, and a kernel function, $k(x, x')$. Given a new input $x_*$, a GP assigns a Gaussian distribution to the output $y_* = y(x^*)$ expressed as:
\begin{align}\label{eq:GPdistribution} 
  p(y_* | x_*, D, \alpha) &= N(y_*| a_*, c^2_* + \sigma^2)   \nonumber \\
  a_* &= k_*^T(K+\sigma^2I)^{-1}y \\
  c^2_* &= k_{**} - k_*^T(K+\sigma^2I)^{-1}k_* \nonumber
\end{align}
where $K$ is a matrix that evaluates the kernel at the training points, $[K]_{ij} = k(x_i, x_j)$, $k_*$ is a vector with $[k_*]_i = k(x_i, x_*)$ and $k_{**}$ is the value of the kernel at $x_*$, $k_{**}= k(x_*, x_*)$. Finally, $y$ represents the vector of observations from the training set, and $\alpha$ is the set of hyperparameters, that includes $\sigma^2$ together with the kernel parameters. These are optimized during the training process. 


In this work we consider the ARD-SE kernel ~\cite{Rasmussen2006} which provides smooth representations of $f(x)$ during GP regression and is the most common kernel employed in the GP-literature. However, it is possible to extend our algorithm to other kernel functions as it is done in \cite{Pan2017}.

\section{GP-SUM Bayes filter}
\label{sec:mixture-filter}

In this section we present GP-SUM, discuss its main assumptions, and describe its computational complexity. Given that GP-SUM is a GP-Bayes filter, our main assumption is that both the dynamics and the measurement models are represented by GPs. This implies that for any state $x_{t-1}$ and action $u_{t-1}$ the probabilities $p(x_t | x_{t-1}, u_{t-1})$ and $p(z_t| x_t)$ are modeled as Gaussian.

To keep track of complex beliefs GP-SUM does not approximate them by single Gaussians, but considers the weaker assumption that they are well approximated by sum of Gaussians. Given this assumption, in \secref{subsec:pred_update} we exploit that the transition and observation models are GPs to correctly propagate the prediction belief, i.e. the pre-observation state distribution. In \secref{subsec:belief_recover} we obtain a close-form solution for the belief expressed as a Gaussian mixture.
\vspace{-0.1in}

\subsection{Updating the prediction belief}\label{subsec:pred_update}

\begin{algorithm}[b]
  \begin{algorithmic}[0]
  \STATE \textbf{GP-SUM}($\{ \mu_{t-1,i}, \Sigma_{t-1,i}, \omega_{t-1,i} \}_{i=1}^{M_{t-1}}$, $u_{t-1}$, $z_{t-1}$, $M_{t}$): \\
  \STATE  $\{ x_{t-1,j} \}_{j=1}^{M_t}$ = sample($\{ \mu_{t-1,i}, \Sigma_{t-1,i}, \omega_{t-1,i} \}_{i=1}^{M_{t-1}}$, $M_t$) \\
    \FOR{$j \in \{1, \dots, M_t\}$ }
      \STATE $\mu_{t,j} = GP_{\mu}(x_{t-1,j}, u_{t-1})$ 
      \STATE $\Sigma_{t,j} = GP_{\Sigma}(x_{t-1,j}, u_{t-1})$ 
      \STATE $\omega_{t,j} = p(z_{t-1} | x_{t-1,j})$ 
    \ENDFOR
    \STATE $\{{\omega}_{t,j}\}_{j=1}^{M_t}$ = normalize$\_$weights($\{{\omega}_{t,j}\}_{j=1}^{M_t}$)
    \STATE \textbf{return } $\{ \mu_{t,j}, \Sigma_{t,j}, \omega_{t,j} \}_{j=1}^{M_{t}}$
  \end{algorithmic}
  \caption{Prediction belief recursion}
  \label{alg:prediction_belief}
\end{algorithm}

The main idea behind GP-SUM is described in \algref{alg:prediction_belief}. Consider \eref{eq:prediction_believe} and \eref{eq:belief_v2}, then the belief at time $t$ in terms of the prediction belief is:
\begin{equation}\label{eq:prediction_belief_v2}
p(x_t | u_{1:t-1}, z_{1:t}) \propto p(z_t| x_t) \cdot p(x_{t} | u_{1:t-1}, z_{1:t-1})
\end{equation}
If the prediction belief at time $t-1$ is approximated by a sum of Gaussians:
\begin{equation}
    \begin{gathered}\label{eq:prediction_believe_t-1} 
p(x_{t-1} | u_{1:t-2}, z_{1:t-2}) = \sum_{i = 1}^{M_{t-1}} \omega_{t-1, i} \cdot \mathcal{N}(x_{t-1}| \mu_{t-1, i}, \Sigma_{t-1, i}) 
\end{gathered}
\end{equation}
where $M_{t-1}$ is the number of components of the Gaussian mixture and $\omega_{t-1,i}$ is the weight associated with the i-th Gaussian of the mixture $\mathcal{N}(x_{t-1}| \mu_{t-1, i}, \Sigma_{t-1, i}) $.

Then we compute the prediction belief at time $t$ combining  \eref{eq:prediction_believe} and \eref{eq:prediction_belief_v2} as:
\begin{equation}
    \begin{gathered}\label{eq:prediction_believe_v3} 
p(x_{t} | u_{1:t-1}, z_{1:t-1}) = \int p(x_t | x_{t-1}, u_{t-1}) p(x_{t-1} | u_{1:t-2}, z_{1:t-1})  d x_{t-1} \propto \\
\int p(x_t | x_{t-1}, u_{t-1}) p(z_{t-1}| x_{t-1}) p(x_{t-1} | u_{1:t-2}, z_{1:t-2}) d x_{t-1} 
\end{gathered}
\end{equation}

Given the previous observation $z_{t-1}$ and the action $u_{t-1}$, the prediction belief at time $t$ can be recursively computed using the prediction belief at time $t-1$ together with the transition and observation models. If $p(x_{t-1} | u_{1:t-2}, z_{1:t-2})$ has the form of a sum of Gaussians, then we can take $M_t$ samples from it, $\{ x_{t-1, j}\}_{j=1}^{M_t}$, and approximate \eref{eq:prediction_believe_v3} by:
\begin{equation}\label{eq:prediction_belief_v4}
p(x_{t} | u_{1:t-1}, z_{1:t-1}) \propto \sum_{j=1}^{M_t} p(x_{t} | x_{t-1,j}, u_{t-1}) p(z_{t-1}| x_{t-1,j})
\end{equation}
Because the dynamics model is a GP,  $p(x_t | x_{t-1,j}, u_{t-1})$ is the Gaussian $\mathcal{N}(x_{t}| \mu_{t, j}, \Sigma_{t, j}) $, and $p(z_{t-1}| x_{t-1,j})$ is a constant value. As a result, we can take:

\begin{equation} \label{eq:weights}
\omega_{t,j} = \frac{p(z_{t-1}| x_{t-1,j})}{\sum_{k=1}^{M_t} p(z_{t-1}| x_{t-1,k})}    
\end{equation}
and express the updated prediction belief again as a Gaussian mixture:
\begin{equation}
p(x_{t} | u_{1:t-1}, z_{1:t-1}) = \sum_{j = 1}^{M_t} \omega_{t, j} \cdot \mathcal{N}(x_{t}| \mu_{t, j}, \Sigma_{t, j}) 
\end{equation}

In the ideal case where $M_t$ tends to infinity, the sum of Gaussians approximation of the prediction belief converges to the real distribution and the propagation over time of the prediction belief remains correct. This property of GP-SUM contrasts with most other GP-Bayes filters where the prediction belief is approximated as a single Gaussian. In those cases, errors from previous approximations inevitably accumulate over time. 

Note that the weights in \eref{eq:weights} are directly related to the likelihood of the observations. As in most sample-based algorithms, if the weights are too small before normalization, it becomes a good strategy to re-sample or modify the number of samples considered. In \secref{sec:results} we address this issue by re-sampling again from the distributions while keeping the number of samples constant. 

\subsection{Recovering the belief from the prediction belief}\label{subsec:belief_recover}


After computing the prediction belief, we take the observation $z_t$ and compute the belief as another sum of Gaussians using \eref{eq:prediction_belief_v2}:
\begin{equation}
\begin{gathered}
p(x_t | u_{1:t-1}, z_{1:t}) \propto p(z_t| x_t)  \sum_{j = 1}^{M_t} \omega_{t, j} \cdot \mathcal{N}(x_{t}| \mu_{t, j}, \Sigma_{t, j}) \\
 = \sum_{j = 1}^{M_t} \omega_{t, j}  \cdot p(z_t| x_t)  \mathcal{N}(x_{t}| \mu_{t, j}, \Sigma_{t, j})  
 \end{gathered}
\end{equation}
Note that if $p(z_t| x_t) \mathcal{N}(x_{t}| \mu_{t, j}, \Sigma_{t, j})$ could be normalized and expressed as a Gaussian distribution, then the belief at time $t$ would directly be a Gaussian mixture. In most cases, however, $p(z_t| x_t)  \mathcal{N}(x_{t}| \mu_{t, j}, \Sigma_{t, j})$ is not proportional to a Gaussian. For those cases, we use standard approximations in the literature (\algref{alg:belief_recovery}). For instance, the algorithm GP-EKF~\cite{Ko2009} linearizes the observation model to express the previous distribution as a Gaussian.

\begin{algorithm}[b]
  \begin{algorithmic}[0]
  \STATE \textbf{belief$\_$computation}($\{ \mu_{t,j}, \Sigma_{t,j}, \omega_{t,j} \}_{j=1}^{M_{t}}$, $z_{t}$, $M_t$): \\
    \FOR{$j \in \{1, \dots, M_t\}$ }
      \STATE $\hat{\mu}_{t,j}, \hat{\Sigma}_{t,j} =$ Gaussian$\_$approx( $p(z_t| x_t) \mathcal{N}(x_{t}| \mu_{t, j}, \Sigma_{t, j})$ )
    \ENDFOR
    \STATE $\{ \hat{\omega}_{t,j}\}_{j=1}^{M_t} = \{{\omega}_{t,j}\}_{j=1}^{M_t}$
    \STATE \textbf{return } $ \{ \hat{\mu}_{t,j}, \hat{\Sigma}_{t,j}, \hat{\omega}_{t,j} \}_{j=1}^{M_{t}} $
  \end{algorithmic}
  \caption{Belief recovery}
  \label{alg:belief_recovery}
\end{algorithm}

In this work, we exploit the technique proposed by  Deisenroth et al.~\cite{Deisenroth2009} as it preserves the first two moments of $p(z_t| x_t) \mathcal{N}(x_{t}| \mu_{t, j}, \Sigma_{t, j})$ and has proven to outperform GP-EKF \cite{Deisenroth2009}. This approximation assumes that $p(x_t,z_t|u_{1:t-1},z_{1:t-1})$ $=p(z_t|x_t) p(x_t|u_{1:t-1},z_{1:t-1})$  and $p(z_t| u_{1:t-1},z_{1:t-1}) = \int p(x_t,z_t|u_{1:t-1},z_{1:t-1}) d x_t $ are both Gaussians. Note that this is an approximation, and that is only true when $x_t$ and $z_t$ are linearly related. Using this assumption and that $p(z_t| x_t)$ is a GP,  $p(z_t| x_t)\mathcal{N}(x_{t}| \mu_{t, j}, \Sigma_{t, j})$ can be approximated as a Gaussian by analytically computing its first two moments~\cite{Deisenroth2009}. As a result, we recover the belief as a sum of Gaussians.

It is key to note that this approximation is only necessary to recover the belief, but does not incur in iterative filtering error. GP-SUM directly keeps track of the prediction belief, for which the approximation is not required.


\subsection{Computational complexity}

The computational complexity of GP-SUM directly depends on the number of Gaussians used at each step. For simplicity, we will now assume that at each time step the number of components is constant, $M$. Note that the number of samples taken from the prediction belief corresponds to the number of components of the next distribution. Propagating the prediction belief one step then requires taking $M$ samples from the previous prediction belief and evaluating $M$ times the dynamics and measurement models. The cost of sampling once a weighted sum of Gaussians is constant, $O(1)$, while evaluating each model implies computing the output of a GP with cost $O(n^2)$, where $n$ is the size of data used to train the GPs \cite{Rasmussen2006}. Therefore the overall cost of propagating the prediction belief is $O(Mn^2 + M)$ where $n$ is the largest size of the training sets considered. Approximating the belief does not represent an increase in $O-$complexity as it also implies $O(Mn^2)$ operations \cite{Deisenroth2009}. 

Consequently GP-SUM's time complexity increases linearly with the size of the Gaussian mixture. When necessary, we can reduce the cost of GP-SUM using sparse GPs~\cite{Pan2017}, which choose a sparser training set, reducing  the cost from $O(Mn^2)$ to $O(Mk^2)$ with $k \ll n$.



\section{Results}
\label{sec:results}

We evaluate the performance of our algorithm in two different dynamic systems. The first one is a 1D synthetic benchmark for nonlinear state space models used in \cite{Deisenroth2009,Kitagawa1996}, where our algorithm proves superior to previous GP-Bayes filters\footnote{The implementations of GP-ADF and GP-UKF are based on \cite{Deisenroth2009} and can be found at \url{https://github.com/ICL-SML/gp-adf}.}. The second case studies how uncertainty propagates in a real robotic system. We learn the stochastic dynamics of planar pushing from real data with GP regression and then use GP-SUM to simulate the system uncertainty overtime to capture the expected distribution of the object state.


\subsection{Synthetic task: algorithm evaluation and comparison}
\label{sec:comparison}

\begin{figure}[t]
\begin{center}
  \includegraphics{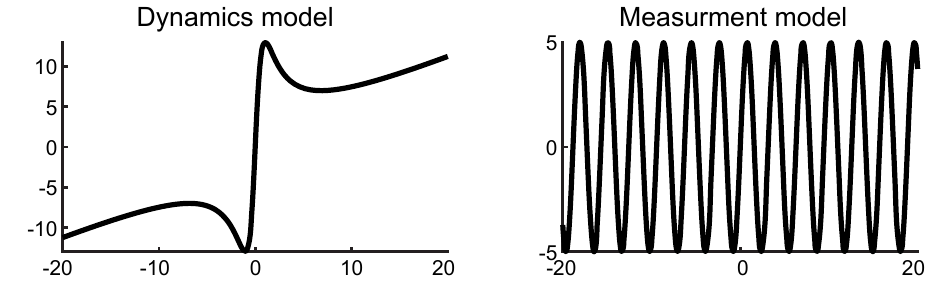}
  \end{center}
  \caption{\textbf{Synthetic benchmark task.} Dynamic model and observation model of the synthetic task in equations \eref{eq:dynamic_model} and \eref{eq:observation_model}. Notice that the dynamics are specially sensitive and non-linear around zero. Just like in the example of the push-grasp in \figref{fig:grasp_mug}, this will lead to unstable behavior and multi-modal state distributions.}
  \label{fig:1D_models}
 \end{figure}

\begin{figure}[!b]
  \begin{center}
\includegraphics{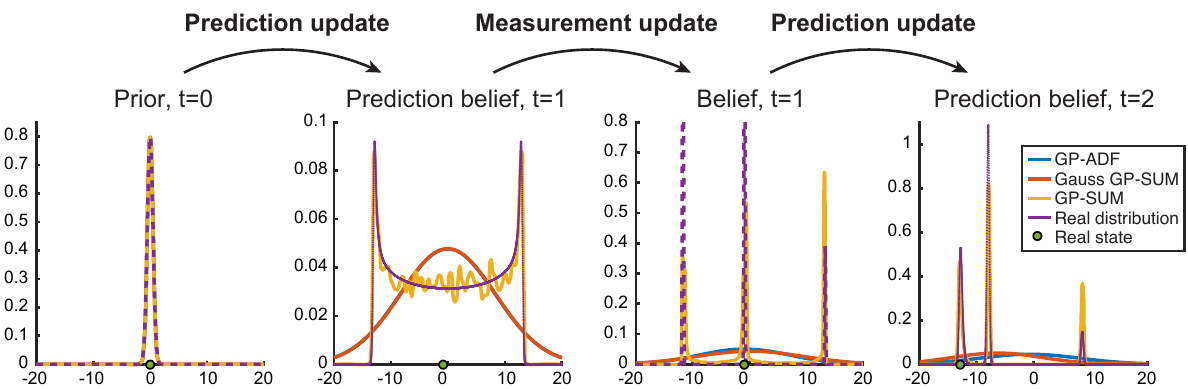}  
\end{center}
  \caption{\textbf{Belief propagation on benchmark synthetic task.} The figure illustrates how GP-SUM, GP-ADF and Gauss GP-SUM propagate the state belief three steps of dynamic-observation-dynamic updates. All three algorithms start from a prior belief centered at zero, precisely where the benchmark dynamic system is most sensitive to initial conditions, as illustrated in \figref{fig:1D_models}. 
  As a result, the belief and prediction belief quickly become multimodal.
  %
  GP-SUM handles properly these complex distributions and its predictions are more accurate. After only one dynamic step, the belief at $t=1$ predicted by GP-SUM shows three possible modes for the state of the system, while the other algorithms output a single Gaussian that encloses them all. }
  \label{fig:1D_example}
\end{figure}

We evaluate GP-SUM on the synthetic system proposed by Kitagawa~\cite{Kitagawa1996}, with dynamics model:
\begin{equation}\label{eq:dynamic_model}
x_{t+1} = \frac{1}{2}x_t + \frac{25 x_t}{1+x_t^2} + w  \qquad w \sim \mathcal{N}(0,0.2^2)    
\end{equation}
and measurement model:
\begin{equation}\label{eq:observation_model}
z_{t+1} = 5 \sin{2x_t} + v \qquad v \sim \mathcal{N}(0,0.01^2)
\end{equation}
The system was previously used to benchmark the performance of GP-ADF~\cite{Deisenroth2009}. \figref{fig:1D_models} illustrates the models and \figref{fig:1D_example} illustrates the filtering process.

The GP models for prediction and measurement are trained using 1000 samples uniformly distributed around the interval $[-20,20]$. GP-SUM uses the same number of Gaussian components $M = M_t = 1000$ during the entire filtering process. The initial prior distribution of $x_0$ is Gaussian with variance $\sigma_0^2 = 0.5^2$ and mean $\mu_0 \in [-10, 10]$. We randomly pick $\mu_0$ 200 times in the interval to assess the filters in multiple scenarios. Their behavior becomes specially interesting around $x=0$ where the dynamics are highly nonlinear. For each value of $\mu_0$, the filters take 10 time-steps. This procedure is repeated 300 times to average the performance of GP-SUM, GP-ADF, GP-UKF, and GP-PF, described in \secref{sec:relatedWork}. For GP-PF, the number of particles is the same as GP-SUM components, $M=1000$.

We evaluate the error in the final state distribution of the system using 3 metrics. The most relevant is the negative log-likelihood, NLL, which measures the likelihood of the true state according to the predicted belief. We also report the root-mean-square error, RMSE, even though it only evaluates the mean of the belief instead of its whole distribution. Similarly, the Mahalanobis distance, Maha, only considers the first two moments of the belief, for which we approximate the belief from GP-SUM by a Gaussian. For the GP-PF we only compute the RMSE given that particle filters do not provide close-form distributions. Note that in all proposed metrics, low values imply better performance.



\begin{table}[t]
  \small
  \caption{Comparison between GP-filters after 1 time step.}
  \label{tab:1D_error_single_step}
  \centering
   \begin{tabularx}{4in}{|p{0.5in}|>{\centering}X|>{\centering}X|>{\centering}X|>{\centering\arraybackslash}X|}
    \hhline{~|----}
    \rowcolor[gray]{.8}
    \multicolumn{1}{c|}{ \cellcolor{white} } & {GP-ADF} & {GP-UKF} & {GP-SUM} &  {GP-PF} \\ \hline
    \rowcolor[gray]{.8}
    Error & 
    $\mu \pm \sigma$ & 
    $\mu \pm \sigma$ & 
    $\mu \pm \sigma$ & 
    $\mu \pm \sigma$ \\ \hline
    NLL &
    \textbf{0.49}  $\pm$ 0.17 & 
    \textbf{95.0} $\pm$ 97.0 & 
    \textbf{-0.55} $\pm$ 0.34  & -\\ \hline
    Maha &
    \textbf{0.69}  $\pm$ 0.06 &
    \textbf{2.80}  $\pm$ 0.72 &  
    \textbf{0.67}  $\pm$ 0.04 & - \\ \hline
    RMSE & 
    \textbf{2.18}  $\pm$ 0.39 & 
    \textbf{34.5} $\pm$ 23.1 & 
    \textbf{2.18}  $\pm$ 0.38 & 
    \textbf{2.27} $\pm$ 0.35\\ \hline
    \end{tabularx}

  \caption{Comparison between GP-filters after 10 time steps.}
  \label{tab:1D_error}
  \centering
   \begin{tabularx}{4.25in}{|p{0.5in}|>{\centering}X|>{\centering}X|>{\centering}X|>{\centering\arraybackslash}X|}
    \hhline{~|----}
    \rowcolor[gray]{.8}
    \multicolumn{1}{c|}{ \cellcolor{white} } &
    {GP-ADF} & {GP-UKF} & {GP-SUM} & {GP-PF}\\ \hline
    \rowcolor[gray]{.8}
    Error &
    $\mu \pm \sigma$ & 
    $\mu \pm \sigma$ & 
    $\mu \pm \sigma$ & 
    $\mu \pm \sigma$ \\ \hline
    NLL &  
    \textbf{9.58} $\pm$ 15.68 & 
    \textbf{1517}  $\pm$ 7600 &
    \textbf{-0.24}  $\pm$ 0.11 & -\\ \hline
    Maha &
    \textbf{0.99} $\pm$ 0.31 & 
    \textbf{8.25} $\pm$ 3.82 &
    \textbf{0.77} $\pm$ 0.06 & -\\ \hline
    RMSE &
    \textbf{2.27} $\pm$ 0.16 & 
    \textbf{13.0} $\pm$ 16.7 & 
    \textbf{0.19} $\pm$ 0.02 & \textbf{ N/A}  \\ \hline
    \end{tabularx}
\end{table}


From \tabref{tab:1D_error_single_step} and \tabref{tab:1D_error}, it is clear that GP-SUM outperforms the other algorithms in all metrics and is more stable, as it obtains the lowest variance in most of the metrics. In the first time step, GP-PF is already outperformed by GP-SUM and GP-ADF, and after a few more time steps, particle starvation becomes a major issue for GP-PF as the likelihood of the observations becomes extremely low. For this reason, we did not report an RMSE value for the GP-PF after 10 time steps. GP-UKF performance is clearly surpassed by GP-SUM and GP-ADF after 1 and 10 time steps.

In \figref{fig:1D_example} we compare the true state distributions (computed numerically) to the distributions obtained by GP-ADF, GP-SUM, and a simplified version of GP-SUM, Gauss GP-SUM, that takes a Gaussian approximation of GP-SUM at each time step. It becomes clear that by allowing non-Gaussian beliefs GP-SUM achieves higher likelihood to the actual state while better approximating the true belief. Instead, GP-ADF's performance is limited by assigning a single Gaussian wide enough to cover all the high density regions. 



\begin{SCfigure}[40][b]
\includegraphics[width=2.3in]{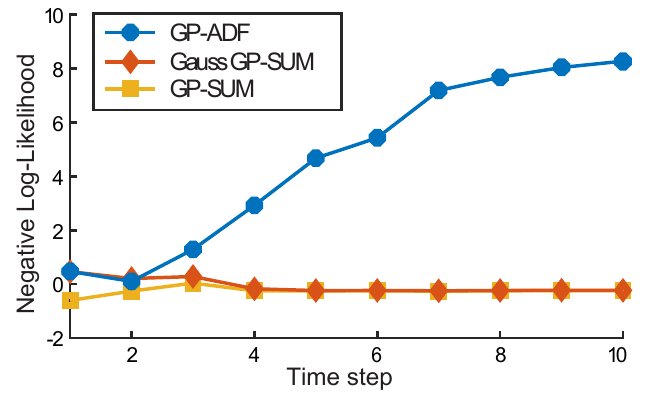}
\caption{\textbf{Filter comparison on benchmark task.} Evaluation of the negative log-likelihood (NLL) of the distributions predicted by filters GP-ADF, Gauss GP-SUM and GP-SUM. In the first time steps, with very few observations, the belief is non-Gaussian or multi-modal. GP-SUM handles this and outperforms both GP-ADF and a Gauss GP-SUM, where at each step the belief is approximated by a Gaussian. As time evolves, both GP-SUM and Gauss GP-SUM converge, since the shape of the real belief becomes uni-modal. GP-ADF worsens with time, since in cases where the dynamics are highly nonlinear, its predicted variance increases, lowering the likelihood of the true state. }
  \label{fig:1D_error}
\end{SCfigure}

\figref{fig:1D_error} shows the temporal evolution of the Negative Log-Likelihood (NLL) metric for GP-ADF, GP-SUM and Gauss GP-SUM.  As the number of steps increases, GP-SUM and Gauss GP-SUM converge because as GP-SUM becomes more confident on the location of the true state, its belief becomes unimodal and more Gaussian. The plot in \figref{fig:1D_error} also shows that GP-ADF worsens its performance over time. This is due to the system crossing states with highly non-linear dynamics, i.e. around zero, where the variance of GP-ADF increases over time. As a result, GP-SUM is a good fit for those systems where multimodality and complex behaviors can not be neglected, at the cost of a larger computational effort.



\subsection{Real task: propagating uncertainty in pushing}
\label{results_pushing}





\begin{figure}[b]
\centering
\includegraphics[width=\linewidth]{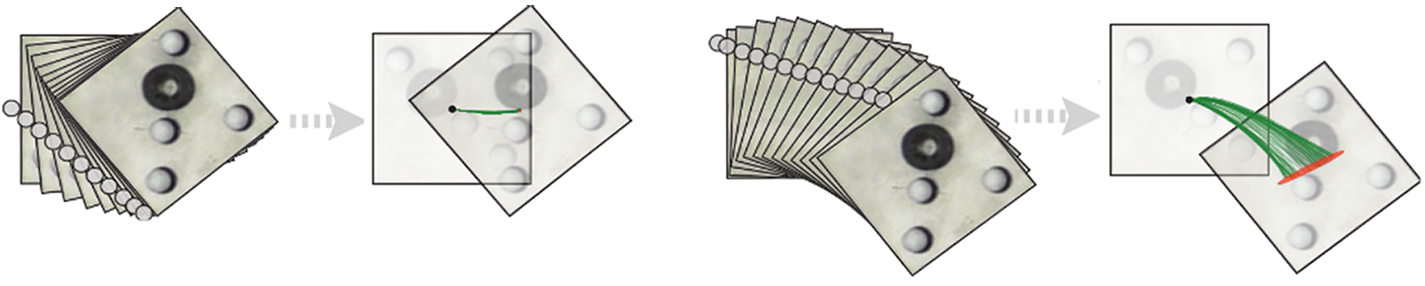}
  \caption{\textbf{Examples of stable and unstable pushes.} Two different pushes whose outcome after 100 executions yields very different distributions: (left) convergent, (right) divergent and multi-modal. Green lines are the trajectories of the geometric center of the object and the orange ellipse approximates the final distribution of the object pose.}
  \label{fig:3_push} 
\end{figure}

Planar pushing is an under-determined and sometimes undecidable physical interaction~\cite{Mason1986}. Only under many assumptions and simplifications can be simulated efficiently~\cite{Goyal1991,Lynch1996}. It has been shown experimentally that due to spatial and direction friction variability, the uncertainty in frictional interactions yields stochastic pushing behavior~\cite{Yu2016,Bauza2017,ma2018friction}.
An important observation is that the type of push has a strong influence in the amount of expected uncertainty, i.e., the level of "noise" in the pushing dynamics is action dependent, a.k.a., heteroscedastic.
\figref{fig:3_push} illustrates this effect, where two different pushes repeated multiple times lead to very different object state distributions.



A stochastic pushing simulation framework that captures heteroscedasticity and non-Gaussian distributions could be relevant for planning and control. We could generate robust plans by preferring those pushes that lead to lower uncertainty and tight distributions for the object's position. To this end, we propose to use GP-SUM to propagate the uncertainty of planar pushing by learning the dynamics of the system using heteroscedastic GPs as proposed in~\cite{Bauza2017}.

%




\begin{figure}[t]
  \begin{center}
    \includegraphics[width=\linewidth]{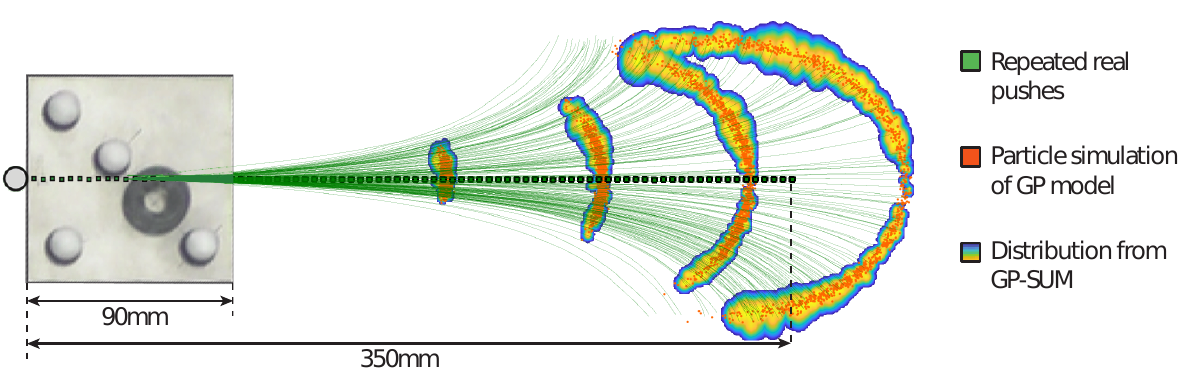}
  \end{center}
    \caption{\textbf{Outcome of an unstable long push.} 
    We execute repeatedly a long push at the center of the side of a square-shaped object. Similar to the push-grasp in \figref{fig:grasp_mug}, the real (green) trajectories show that the block can rotate to either side of the pushing trajectory---it is naturally unstable and undecidable~\cite{Mason1986}.
    The stochastic GP model from~\cite{Bauza2017} can capture that uncertainty in the form of a probabilistic distribution. The (orange) dots show the outcome of $1000$ Monte Carlo simulations of the learned GP dynamic model. GP-SUM predicts accurately the ring-shaped distribution in a way that is not possible with standard GP-filters that assume a uni-modal Gaussian form for the state belief.}
    \label{fig:push_dist}
\end{figure}

In this case, since simulation is only concerned about forward propagation of uncertainty, we do not consider a measurement model. As a result, when using GP-SUM the prediction belief and the belief coincide and all the components in the Gaussian mixtures have the same weights. In the absence of an observation model, the simulated state distributions naturally become wider at  step. For the following results we use $M=10000$ Gaussians to capture with accuracy the complex distributions originated in long pushes.

We train the GP model of the dynamics with real data and take as inputs the contact point between the pusher and the object, the pusher's velocity, and its direction~\cite{Bauza2017}. The outputs of the dynamics model are the displacement---position and orientation---of the object relative to the pusher's motion. Each real push for validation is repeated 100 times at a velocity of $20$mm/s. The intrinsic noise of the system, combining the positional accuracy of the robot and the positional accuracy of the object tracking system, has a standard deviation lower than $1$mm over the object location and lower than $0.01$rad for the object orientation.

Since the distribution of the object position easily becomes non-Gaussian, GP-SUM obtains more accurate results than other algorithms. \figref{fig:push_dist} shows an example of a $350$mm long push at the center of one of the sides of a squared object. We compare the real pushing trajectories (green) with the outcome of running a montecarlo particle simulation on the learned GP-model and GP-SUM's prediction. The distribution becomes ring-shaped and multi-modal, which GP-SUM has no trouble in recovering. This property cannot be captured by standard GP-Bayes filters that assume single Gaussian state distributions. 
 

\begin{figure}[t]
\centering
\includegraphics[width=3in]{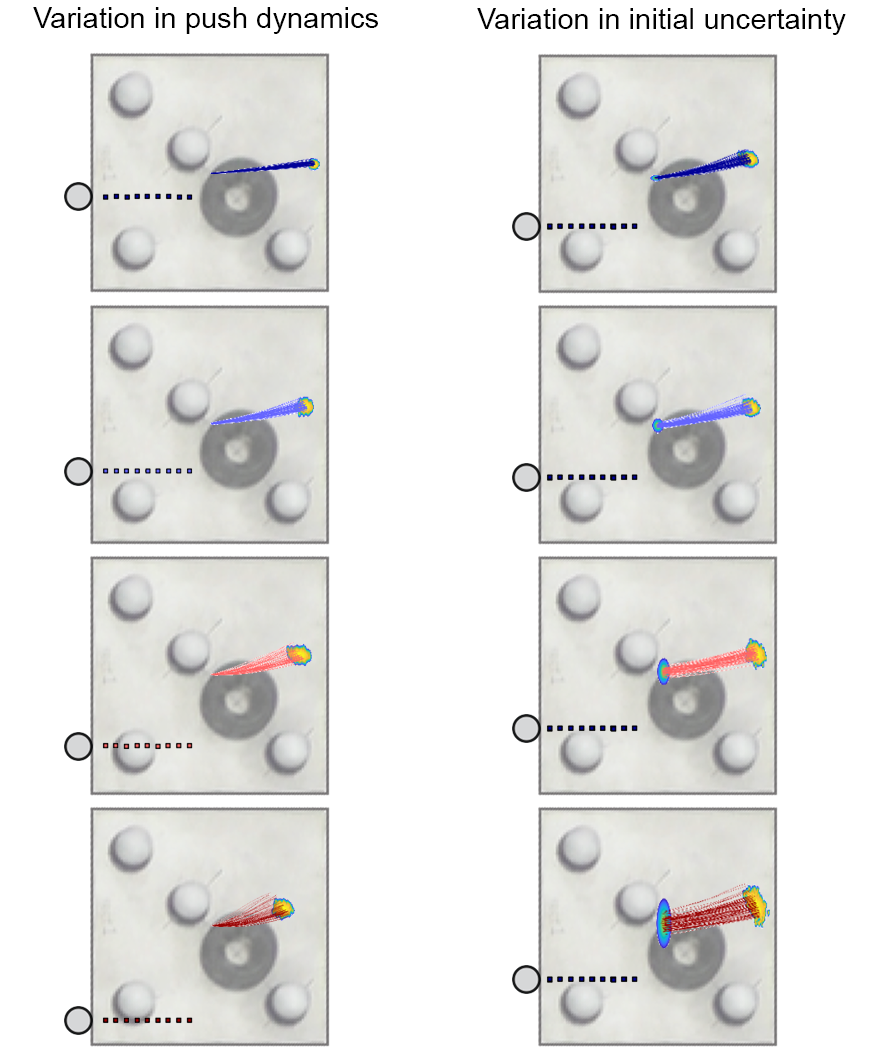}
  \caption{\textbf{GP-SUM capturing variations in output uncertainty.} [Left] Four different straight pushes of $5$cm, and the resulting distributions of the location of the geometric center of the object. We compare the outcome of  100 repeated pushes with GP-SUM's prediction. [Right] Similar analysis, but changing the amount of uncertainty in the initial location of the geometric center of the object (Gaussian distribution with $ \sigma=\{0, 0.5, 1, 2\}$mm added on top of the noise of the Vicon tracking system). In both cases, GP-SUM successfully approximates the shape of the output distributions.}
\label{fig:diff_pushes}
\label{fig:push_types}
\label{fig:push_initial_dist}
\end{figure}
Being able to propagate the uncertainty of the object position over time exposes interesting properties of the planar pushing system. For instance in \figref{fig:push_types} we observe different pushes repeated many times and how GP-SUM can obtain a reasonable approximation of the true distribution and recover the different amounts of noise produced by each type of push, i.e., the heteroscedasticity of the system. \figref{fig:push_initial_dist} also shows how GP-SUM can take into account different initial noise distributions and propagate properly the uncertainty in the object's position. Being able to recover these behaviors is specially useful when our goal is to push an object to a specific region of the space as it allows to distinguish between pushes that lead to narrower (low-variance) distributions and those that involve multimodal or wider (high-variance) distributions. 



\section{Discussion and Future work}
\label{sec:discussion}

GP-Bayes filters are a powerful tool to model and track systems with complex and noisy dynamics. Most approaches rely on the assumption of a Gaussian belief.
This assumption is an effective simplification. It enables filtering with high frequency updates or in high dimensional systems. It is most reasonable in systems where the local dynamics are simple, i.e., linearizable, and when accurate observation models are readily available to continuously correct for complex or un-modelled dynamics.

In this paper we look at situations where the Gaussian belief is less reasonable. That is, for example, the case of contact behavior with non-smooth local dynamics due to sudden changes in stick/slip or contact/separation, and is the case in stochastic simulation where, without the benefit of sensor feedback,  uncertainty distributions naturally grow over time.
We propose the GP-SUM algorithm which exploits the synergies between dynamic models expressed as GPs, and complex state distributions expressed as a weighted sum of Gaussian.

Our approach is sample-based in nature, but has the advantage of using a minimal number of assumptions compared to other GP-filters based on single Gaussian distributions or the linearization of the GP-models. Since GP-SUM preserves the probabilistic nature of a Bayes filter, it also makes a more effective use of sampling than particle filters.

When considering GP-SUM, several aspects must be taken into account:

\myparagraph{Number of samples.} Choosing the appropriate number of samples determines the number of Gaussians in the prediction belief and hence its expressiveness. Adjusting the number of Gaussians over time is likely beneficial in order to properly cover the state space. Similarly, high-dimensional states might require higher values of $M_t$ to ensure a proper sampling of the prediction belief. Because of the sample-based nature of GP-SUM, many techniques from sample-based algorithms can be effectively applied such as resampling or adding randomly generated components to avoid particle deprivation.

\myparagraph{Likelihood of the observations.} There is a direct relation between the weights of the beliefs and the likelihood of the observations. We can exploit this relationship to detect when the weight of the samples degenerates and correct it by re-sampling or modifying the number of samples. 

\myparagraph{Computational cost.} Unlike non-sampling GP-filters, the cost of GP-SUM scales linearly with the number of samples. Nevertheless, for non-linear systems we showed that our algorithm can recover the true state distributions more accurately and thus obtain better results when compared to faster algorithms such as GP-ADF, GP-UKF or GP-PF. 

\myparagraph{GP extensions.} The structure of GP-SUM is not restricted to classical GPs for the dynamics and observation model. Other types of GPs such as HGPs or sparse GPs can be considered. For instance, combining GP-SUM with SSGPs \cite{Pan2017} makes the computation more efficient.

Future research will focus on combining GP-SUM with planning and control techniques. Simulating multimodality and noisy actions can provide a guidance to chose actions that better deal with the dynamics of complex stochastic systems.
\vspace{-3mm}
%

\addcontentsline{toc}{section}{References}
{\tiny \bibliographystyle{include/splncs}}
\bibliography{references}

\end{document}